%% file: main.tex
\documentclass[letterpaper, 10 pt, conference]{ieeeconf}

\IEEEoverridecommandlockouts                              

\usepackage{cite}
\usepackage{graphics} 
\usepackage{epsfig} 
\usepackage{times} 
\usepackage{amsmath} 
\usepackage{amssymb}  
\usepackage{subfig}
\usepackage{xcolor} 
\usepackage{color,soul}
\usepackage{graphicx}
\usepackage{xy}
\usepackage{atbegshi}

\usepackage{import,bm}

\title{Learning Therapist Policy from Therapist-Exoskeleton-Patient Interaction}

\author{Grayson Snyder, Lorenzo Vianello, Levi Hargrove, Matthew L. Elwin, Jose Pons  
\thanks{Corresponding author: lvianello@sralab.org}
\thanks{G. Snyder, M. Elwin are with Center for Robotics and Biosystems, Northwestern University, Evanston, IL, USA.}
\thanks{L. Vianello, L. Hargrove, and J. Pons are with the Legs and Walking Lab, Shirley Ryan AbilityLab, Chicago, IL, USA.}
}

\begin{document}

\maketitle
\thispagestyle{empty}
\pagestyle{empty}


\begin{abstract}
\begin{color}{black}
Post-stroke rehabilitation is often necessary for patients to regain proper walking gait. However, the typical therapy process can be exhausting and physically demanding for therapists, potentially reducing therapy intensity, duration, and consistency over time.
\end{color}
We propose a Patient-Therapist Force Field (PTFF) to visualize therapist responses to patient kinematics and a Synthetic Therapist (ST) machine learning model to support the therapist in dyadic robot-mediated physical interaction therapy. 
\begin{color}{black}
The first encodes patient and therapist stride kinematics into a shared low-dimensional latent manifold using a Variational Autoencoder (VAE) and models their interaction through a Gaussian Mixture Model (GMM), which learns a probabilistic vector field mapping patient latent states to therapist responses. This representation visualizes patient–therapist interaction dynamics to inform therapy strategies and robot controller design.
The latter is implemented as a Long Short-Term Memory (LSTM) network trained on patient–therapist interaction data to predict therapist-applied joint torques from patient kinematics. Trained and validated using leave-one-out cross-validation across eight post-stroke patients, the model was integrated into a ROS-based exoskeleton controller to generate real-time torque assistance based on predicted therapist responses.
\end{color}
Offline results and preliminary testing indicate the potential of their use as an alternative \begin{color}{black} approach to\end{color} post-stroke exoskeleton therapy. The PTFF provides understanding of the therapist's actions while the ST frees the human therapist from the exoskeleton, allowing them to continuously monitor the patient's nuanced condition. 
\end{abstract}

\section{Introduction}

\input{sections/01_Intro}

\begin{figure*}[t]
    \vspace{0.3cm}
    \centering
    \includegraphics[width=0.9\linewidth]{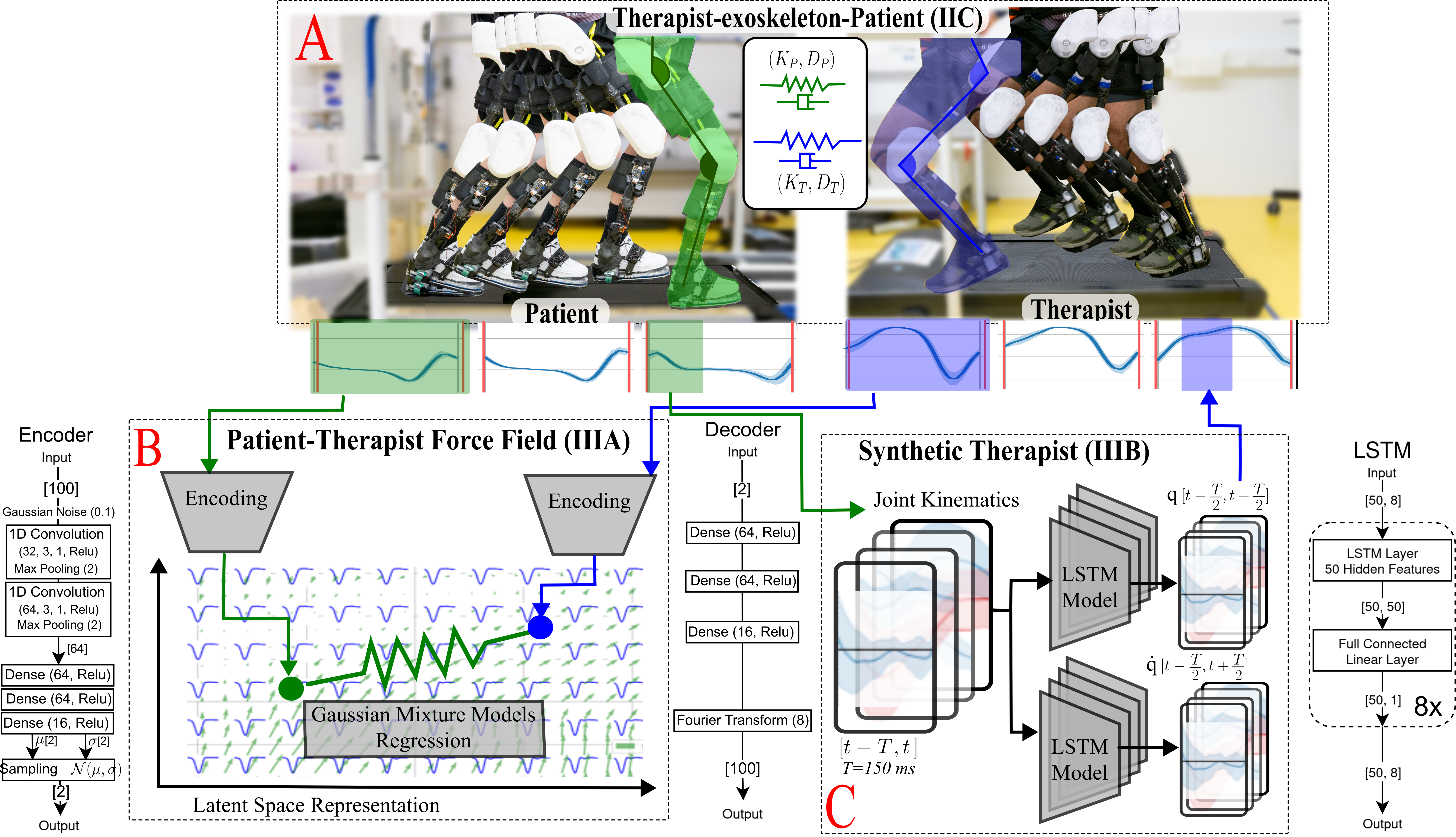}
    \caption{
    \begin{color}{black}
    \small{Learning Therapist Policy from Therapist–Exoskeleton–Patient Interaction. (A) Previously collected dataset of therapist–exoskeleton–patient physical interaction through two lower-limb exoskeletons. The exoskeletons render virtual springs connecting the joint configurations of the patient and therapist. (B) Patient–Therapist Force Field, which takes as input the patient’s and therapist’s strides, extracts latent representations, and learns a force field connecting them in a lower-dimensional space. The model uses an encoder–decoder architecture (both displayed) to reduce dimensionality.
    (C) Synthetic Therapist, which takes as input the patient’s joint kinematics and predicts the corresponding therapist kinematics. The model uses a Long Short-Term Memory (LSTM) network (displayed) to infer the therapist’s trajectory.}
\end{color}
}
    \label{fig:initialImg}
    \vspace{-0.3cm}
\end{figure*}

\section{Background}
\label{sec:background}

\input{sections/02_SoA_LfD}

\input{sections/03_DataSetPresentation}


\section{Methods}
\label{sec:methods}

This section introduces the two frameworks presented in this work, namely the \textit{Patient-Therapist Force Field} and the \textit{Synthetic Therapist}.

\input{sections/04_ForceFieldMethods}

\begin{figure*}[t]
    \vspace{0.2cm}
    \centering
    
    \includegraphics[width=0.7\linewidth]{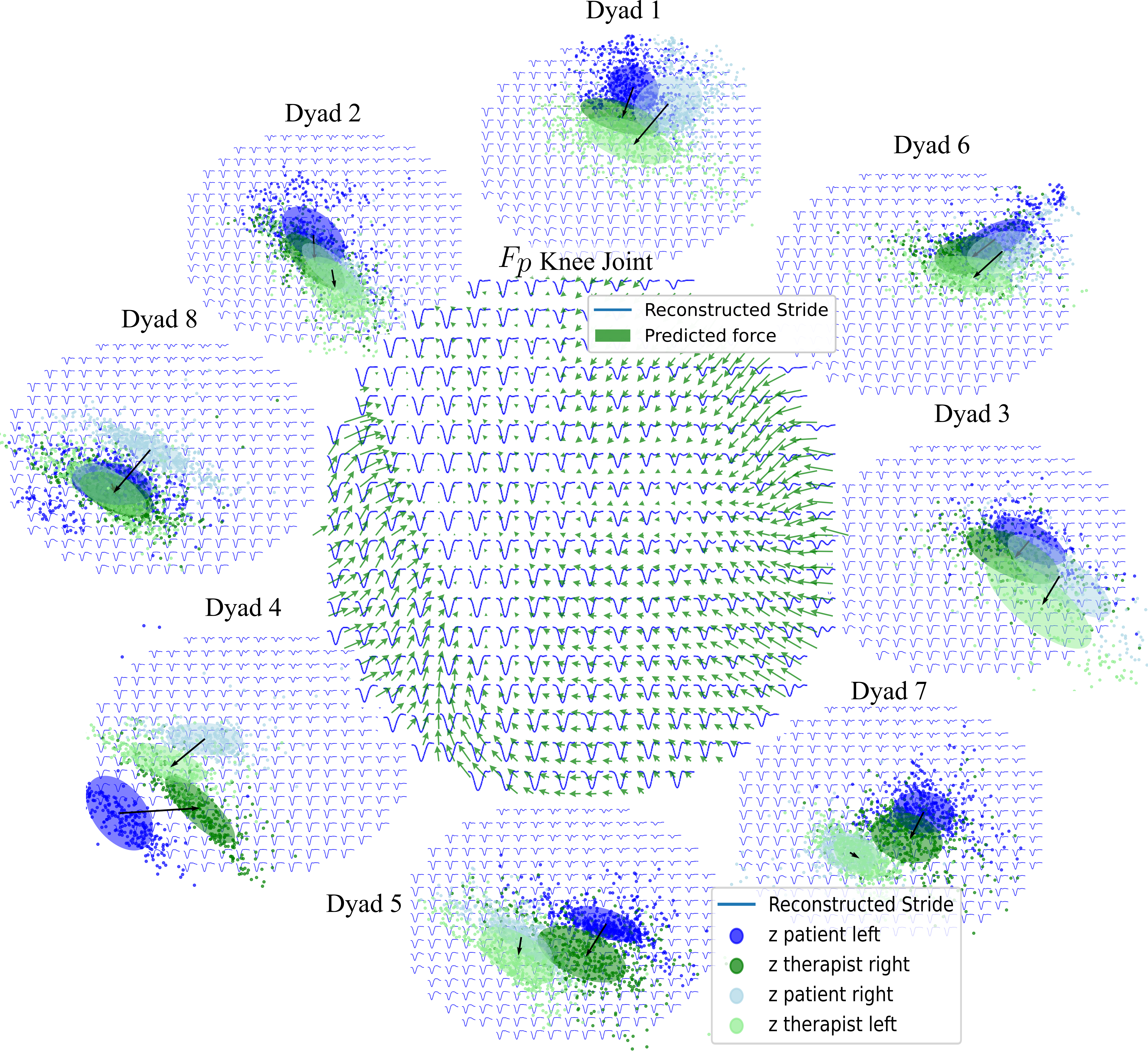}
    
    \vspace{0.3cm}
    
    \includegraphics[width=0.85\linewidth]{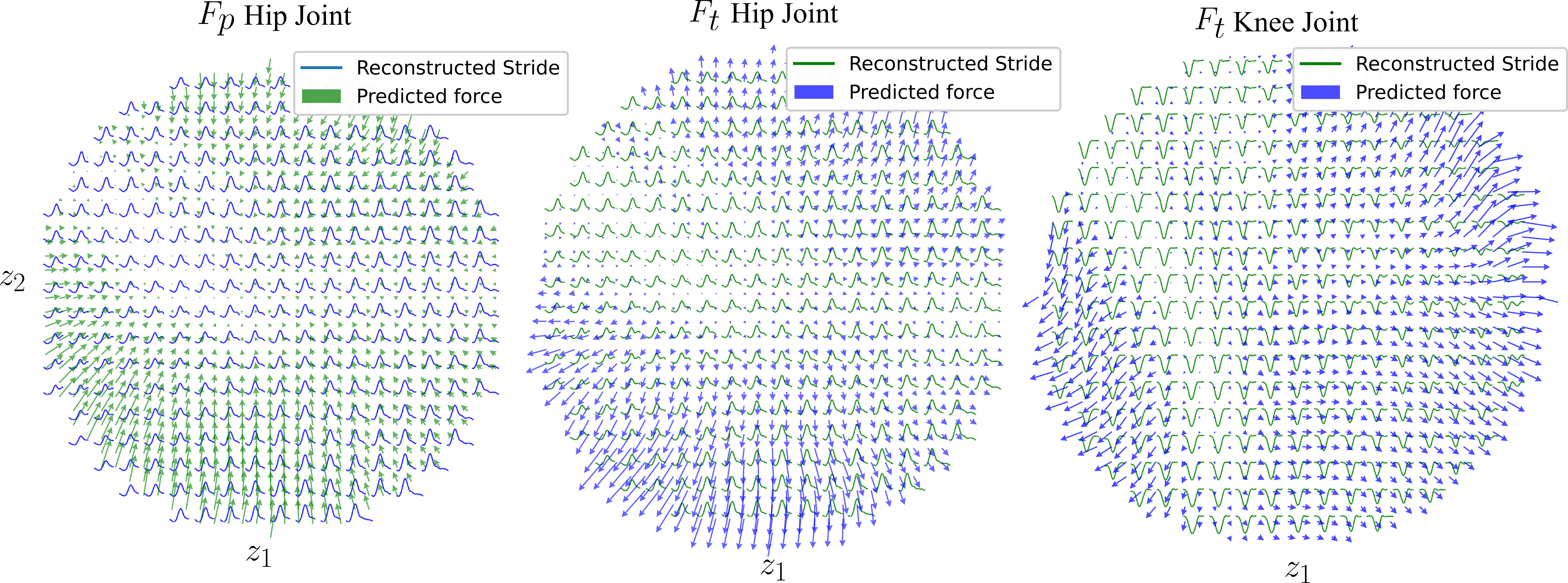}
    
    \caption{\small{
    \begin{color}{black}
    \textbf{Top:} Latent space representation of each individual dyad. Each dot corresponds to a stride, while the ellipsoids display their distribution. Blue: patient left stride; green: therapist right stride; light blue: patient right stride; light green: therapist left stride. The connection is mirrored (patient left, therapist right). In the middle the resulting force field computed by fitting all the dyads using the GMM.
    \textbf{Bottom:} Force fields (from left to right): patient hip, therapist hip, patient knee, therapist knee.
    \end{color}
    }}
    
    \label{fig:LatentAndForceFields}
    \vspace{-0.3cm}
\end{figure*}

\input{sections/05_SyntheticTherapistMethods}

\section{Results \& Discussion}
\label{sec:results}

\begin{table}[t]
\vspace{0.4cm}
\center
\begin{tabular}{cc|cc|}
\cline{3-4}
                                &          & \multicolumn{2}{c|}{rMSE (deg, deg/s)}                          \\ \hline
\multicolumn{1}{|c|}{Train/Validation-Set} & \multicolumn{1}{c|}{Test-Set} & \multicolumn{1}{c|}{Position Avg.} & \multicolumn{1}{c|}{Velocity Avg.} \\ \hline
\multicolumn{1}{|c|}{PT$_{1-8}$}    & \multicolumn{1}{c|}{PT$_{1-8}$}   & \multicolumn{1}{c|}{$4.30 \pm 0.41$}           & \multicolumn{1}{c|}{$20.67 \pm 1.47$}           \\ \hline
\multicolumn{1}{|c|}{PT$_{2-8}$}    & \multicolumn{1}{c|}{PT$_{1}$}   & \multicolumn{1}{c|}{$5.93 \pm 1.20$}           & \multicolumn{1}{c|}{$25.13 \pm 4.54$}           \\ \hline
\multicolumn{1}{|c|}{PT$_{1,3-8}$}    & \multicolumn{1}{c|}{PT$_{2}$}   & \multicolumn{1}{c|}{$5.77 \pm 1.12$}           & \multicolumn{1}{c|}{$25.47 \pm 5.03$}           \\ \hline
\multicolumn{1}{|c|}{PT$_{1,2,4-8}$}    & \multicolumn{1}{c|}{PT$_{3}$}   & \multicolumn{1}{c|}{$6.48 \pm 0.67$}           & \multicolumn{1}{c|}{$33.62 \pm 3.87$}           \\ \hline
\multicolumn{1}{|c|}{PT$_{1-3,5-8}$}    & \multicolumn{1}{c|}{PT$_{4}$}   & \multicolumn{1}{c|}{$4.95 \pm 1.00$}           & \multicolumn{1}{c|}{$20.66 \pm 1.70$}           \\ \hline
\multicolumn{1}{|c|}{PT$_{1-4,6-8}$}    & \multicolumn{1}{c|}{PT$_{5}$}   & \multicolumn{1}{c|}{$3.76 \pm 0.15$}           & \multicolumn{1}{c|}{$18.27 \pm 1.08$}           \\ \hline
\multicolumn{1}{|c|}{PT$_{1-5,7,8}$}    & \multicolumn{1}{c|}{PT$_{6}$}   & \multicolumn{1}{c|}{$5.07 \pm 1.22$}           & \multicolumn{1}{c|}{$23.38 \pm 2.06$}           \\ \hline
\multicolumn{1}{|c|}{PT$_{1-6,8}$}    & \multicolumn{1}{c|}{PT$_{7}$}   & \multicolumn{1}{c|}{$4.43 \pm 0.98$}           & \multicolumn{1}{c|}{$19.51 \pm 1.21$}           \\ \hline
\multicolumn{1}{|c|}{PT$_{1-7}$}    & \multicolumn{1}{c|}{PT$_{8}$}   & \multicolumn{1}{c|}{$4.00 \pm 0.43$}           & \multicolumn{1}{c|}{$19.51 \pm 1.61$}           \\ \hline
\end{tabular}
\caption{\small Prediction accuracy expressed as mean root Mean-Squared-Error (rMSE) for the LSTM model joint position and joint velocity prediction. We performed a cross-user generalization evaluation by iteratively training on all users (PT$_i$) except one and testing on the excluded user.}
\label{tab:leaveOneOut}
\vspace{-0.3cm}
\end{table}

\subsection{Patient-Therapist Force field}

\input{sections/06_ForceFieldResults}

\begin{figure}[t]
    \vspace{0.4cm}
    \centering
    \includegraphics[width=0.9\linewidth]{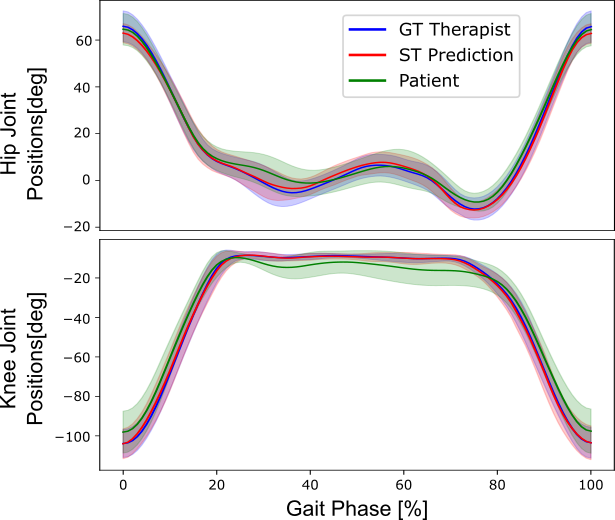}
    \caption{\small Example of Synthetic Therapist predictions (predicting 75ms into the future) overlayed with corresponding true therapist data and patient data. On top is presented the Hip Joint, while on the bottom the Knee joint. }
    \label{fig:PredVsTrue}
    \vspace{-0.3cm}
\end{figure}

\subsection{Synthetic Therapist}

\input{sections/07_SynthticTherapistResults}

\section{Conclusion}
\input{sections/08_Conclusions}

\bibliographystyle{IEEEtran}
\bibliography{sample,references}

\end{document}

%% file: sections/01_Intro.tex
In physical therapy, motor-impaired patients often depend on direct physical assistance from therapists to restore mobility and improve motor function \cite{rudberg2021stroke}. 
Current interventions primarily focus on high-intensity gait training administered by physical therapists \cite{belda2011rehabilitation}. For example, during post-stroke gait rehabilitation, therapists frequently guide foot placement to correct abnormal walking patterns. 

This approach demands significant physical effort from therapists, which can cause work-related injuries \cite{mccrory2014work} and limit their ability to provide full-body guidance and assessments. \begin{color}{black}Due to the need to interact at multiple contact points with the patient during full-body interaction,\end{color} training often requires multiple therapists to coordinate interventions across several joints simultaneously \cite{stephan2021mobility}.

Lower-limb exoskeletons can support intensive multi-joint gait rehabilitation, reducing therapist effort, sustaining patient weight, and providing objective measures of performance~\cite{belda2011rehabilitation, moeller2023use, gassert2018rehabilitation}. Common control strategies include assist-as-needed, which guides patients along predefined trajectories to support learning~\cite{hobbs2020review, baud2021review, belda2011rehabilitation}, and error augmentation, which introduces resistance or deviations to promote adaptation~\cite{marchal2019haptic, Marchal-Crespo2009}.

\begin{color}{black}
Nevertheless, defining optimal reference trajectories and tuning the appropriate level of assistance or resistance in exoskeleton therapy remain open challenges~\cite{belda2011rehabilitation, zhang2017human}. In practice, many systems rely on predefined trajectories derived from able-bodied walking, which fail to adapt to individual patient impairments. Moreover, fixed assistance levels may not reflect the patient’s evolving motor capacity, and current control strategies often underutilize the therapist’s expertise in providing nuanced, real-time adjustments~\cite{hasson2023neurorehabilitation}. As a result, therapists have limited ability to modulate guidance during therapy, potentially reducing functional benefits and hindering the clinical adoption of exoskeleton-based rehabilitation~\cite{baud2021review, celian2021day}.
\end{color}

Physical Human-Robot-Human Interaction (pHRHI) uses robotic devices to mediate physical interaction between humans, merging human motor-cognitive skills with robotic capabilities~\cite{Kucuktabak2021, vianello2025robot}. In pHRHI, two users interface with separate robots connected via virtual springs and dampers, allowing each to feel and respond to their partner’s movements.
\begin{color}{black}
Studies in both upper-limb~\cite{Beckers2020, Kager2019TheTask} and lower-limb~\cite{ Short2023, short2025effects, Kucuktabak2023, Vianello2024, koh2021exploiting} tasks have shown promising results in healthy populations, demonstrating improved motor performance and increased engagement. More recent studies have also reported promising outcomes in post-stroke populations for upper-limb tasks~\cite{waters2024theradyad}. In our previous work, we validated this approach for lower-limb rehabilitation by virtually connecting two exoskeletons to enable interaction between a therapist and eight post-stroke patients~\cite{kuccuktabak2025therapist}.
\end{color}

Thus, by delving into previously collected data of patients and a therapist virtually connected during a clinically relevant rehabilitation exercise, we have developed a preliminary approach that uses pHRHI interactions to capture therapist expertise with machine learning techniques capable of (1) identifying therapist corrective strategies and (2) reproducing these actions to support or partially substitute the therapist during extended training sessions.

The contribution of this work is two-fold: (1) we introduce \textit{Patient-Therapist Force Field} (PTFF), a novel approach to visualize a therapist's response to patient kinematics, allowing better understanding of the therapist's actions and decision making during therapy;  (2) a \textit{Synthetic Therapist} (ST) that learns from patient-therapist demonstrations and infers therapist response based on a short history of patient kinematics. Both approaches have been validated
on previously collected data from eight post-stroke patients, and have potential to enable better understanding of a therapist's actions during therapy while freeing the therapist from the exoskeleton. Fig. \ref{fig:initialImg} displays the overall infrastructure presented in this work. 

The paper is structured as follows: in Sec.~\ref{sec:background}, we define the concept of the therapist’s policy, review preliminary studies that attempt to identify and reconstruct such a policy, and present a holistic previously-collected dataset of eight post-stroke patients to support policy creation. Sec.~\ref{sec:methods} describes the two proposed approaches, namely PTFF and ST. The paper concludes with the validation of the results and discussion of the two approaches in Sec.~\ref{sec:results}.

%% file: sections/02_SoA_LfD.tex
\begin{color}{black} In this section, we discuss issues in the creation of a therapist policy (Sec. \ref{sec:therapist_policy}) and previous attempts to understand and model the therapist policy ($\pi(.)$)(Sec. \ref{sec:SoA}), while in the following section (Sec. \ref{sec:dataset}), we present preliminary data that potentially allows us to encode all the discussed information. \end{color}

\subsection{Therapist Policy}
\label{sec:therapist_policy}
Machine Learning has been extensively used to control lower-limb exoskeletons \cite{coser2024ai}. Among these, methods like \cite{molinaro2024task} regress user assistance level across different activities, providing assistance to both healthy and patient populations. However, defining optimal control strategies during a rehabilitation session is a more differentiated task because each patient has specific needs that the therapist must identify at each session \cite{poggensee2021adaptation, lhoste2025benchmark}. For instance, therapists often choose to assist one joint and resist another one. 
We define the \textit{Therapist Policy} as a function $\pi:\mathcal{S}_p \rightarrow \mathcal{A}_t$ that, given the current patient state $s_p \in S_p$, returns a therapist action $a_t \in \mathcal{A}_t \subset \mathbb{R}^m$. The understanding and subsequent design of a policy of this kind is still an open research question. 

Several barriers prevent a systematic understanding of the therapist’s choices during interaction with a patient. One challenge is the design of patient state space $(\mathcal{S}_p \subset \mathbb{R}^n)$ that includes multiple features, namely the impairment level and location, the resulting kinematics, kinetics, and dynamics, or even emotional and physiological aspects. These aspects result in a high dimensionality representation; i.e., a state space with large $n$. Similarly, it is difficult to design a therapist action space $(\mathcal{A}_t \subset \mathbb{R}^m)$ that encodes the multichannel actions the therapist can perform (physically, visually, and vocally) that allows the overall system to be controllable. In other words, what actions enable the therapist to drive the patient to an optimal state $s_p^*$? Another challenge is the need for high measurability of the patient and the therapist. Kinematics information can be easily extracted from camera or inertial systems \cite{bartloff2025advancing} while kinetics information requires more burdened measures coming from electromyography \cite{munoz2022upper}. Measuring emotional information, dynamics, and precise forces exchanged between therapist and patient present additional difficulties\cite{galvez2005measuring}. 

\subsection{Related Work}
\label{sec:SoA}

Luciani et al. \cite{luciani2024imitation, luciani2024therapists} introduced a Learning by Demonstration (LbD) framework that first estimates a therapist's force contribution while manipulating a patient's arm with an upper-limb exoskeleton. The data is then used to learn an interaction model between patient and therapist.

Similarly, in \cite{sankar2024action} LbD is used to train a model based on a diffusion policy, using data from a therapist–patient dyad connected to lower-limb exoskeletons. Their machine learning model aimed to replicate therapist policies even in the absence of the therapist, but struggled with overfitting due to reliance on data from a single patient and high inference time. Using a different style of model and data collected from more patients, we aim to decrease inference time to real-time and create a more generalizable system for use across varying patients.

Moreover, all previous approaches perform an end-to-end regression of the therapist's action, missing an in-depth analysis of the mechanistic properties of the physical interaction between the two actors. This path, while more time-consuming, can promote safety by avoiding unexpected and dangerous therapist actions, help with debugging, and help unmask the motor learning mechanism. 
A first step in this direction is a minimal representation of the patient state ($s_p$). Shushtari \cite{shushtari2024humanexoskeleton} introduces the Interaction Portrait (IP), which can help visualize Human-Exoskeleton interactions in polar coordinates. This approach allows for visualizing a patient's dynamic behavior in an understandable two-dimensional coordinate system, but at the same \begin{color}{black}time\end{color} it lacks a clear visualization of the therapist's response to that behavior.

%% file: sections/03_DataSetPresentation.tex
\subsection{Therapist-exoskeleton-Patient Dataset}
\label{sec:dataset}

\begin{color}{black}
In our previous study \cite{kuccuktabak2025therapist} two lower-limb exoskeletons (X2, Fourier Intelligence), shown in Figure~\ref{fig:initialImg}, were adapted to implement the robot-mediated physical interaction framework for gait therapy. 
\end{color}
Each exoskeleton renders a virtual interaction medium modeled as torsional spring–damper elements between the hip and knee joints of therapist and patient. These elements, adjustable for each user, define the level of haptic feedback. During training, the therapist and patient face each other, with their opposite legs virtually coupled (therapist’s left/right with patient’s right/left). Using the interaction element properties and instantaneous joint states ($t$: therapist, $p$: patient), the desired interaction torques for each joint were computed as
\begin{align}
     \hat{\tau}_t^* &= K_t(\tilde{\theta}_p - \theta_t) + B_t(\dot{\tilde{\theta}}_p - \dot{\theta}_t), \\
     \hat{\tau}_p^* &= K_p(\tilde{\theta}_t - \theta_p) + B_p(\dot{\tilde{\theta}}_t - \dot{\theta}_p),
     \label{eq:joint_space}
\end{align}
where $\hat{\tau}_i^*:i \in {t, p}$ are the desired interaction torques for the exoskeleton of the therapist and the patient, $(K_i, B_i): i \in {t, p}$ are the stiffness and damping components, $(\theta_i, \dot{\theta}_i)$ and  $(\tilde{\theta}_i, \dot{\tilde{\theta}}_i):i \in {t, p}$ are the joint angles and velocities of the two opposite legs. Desired interaction torques for each joint of each user were commanded to their corresponding exoskeleton as in ~\cite{Kucuktabak2023}.

The open-source dataset from our previous work \cite{kuccuktabak2025therapist} includes \begin{color}{black} 
eight patients with chronic hemiparetic stroke ( $> 6$ months post-onset), all trained by the same licensed physical therapist. Each dyad (therapist and patient) performed three sessions consisted of 10-minute training with ~3-minute breaks.
\end{color}
During training, the therapist assisted via the virtual exoskeleton connection and auditory cues. Between blocks, the therapist could adjust haptic feedback (i.e., $K_t$ in Eq. \ref{eq:joint_space}) to ease control. Exoskeleton kinematics were recorded at 333 Hz via hip and knee encoders, while an IMU on the backpack measured trunk orientation.

%% file: sections/04_ForceFieldMethods.tex
\subsection{Patient-Therapist Force Field}

The PTFF enables visualization of the stride execution of both the patient and the therapist in a two-dimensional space. The method relies on two consecutive mechanisms: (1) a Variational Autoencoder (VAE) to reduce the dimensionality of the patient’s and therapist’s strides, and (2) a Gaussian Mixture Model (GMM) to learn the force field that describes the relationship between the low-dimensional latent representations of the patient and the therapist. In this section, we first describe how the dataset previously collected \begin{color}{black}(Sec. \ref{sec:dataset})\end{color} was processed, how the VAE was trained and used to extract the latent representation and the latent space manifold. Finally, we discuss the training and validation of the proposed force field.

\subsubsection{Data Preparation}
The data were processed by synchronizing patient and therapist kinematics ($q_T, q_P$). All strides were segmented relative to the patient’s heel-strike, identified through foot plate sensors detecting ground contact. Left and right strides were segmented separately for each leg, with each stride composed by both the kinematic motion of the hip and the knee joint $s = (s_h, s_k) \in \mathbb{R}^{(2,100)}$ starting and ending at heel-strike. Outlier strides, defined as those outside the 90th percentile, were removed. Left and right strides were then combined into a single dataset $S \subset \mathbb{R}^{(2,100)}$, yielding a total of 12K patient strides with the corresponding therapist strides. For each stride, several additional features were stored to enable post-processing, including stride ownership (patient or therapist), patient ID, leg side (left or right), the correspondence between patient and therapist strides ($s^p \leftrightarrow s^t$), and the stiffness values applied to both patient and therapist ($K_p, K_t$). 
After preprocessing, all strides were normalized using mean and standard deviation and split into training and validation sets with a 70/30 ratio. 

\subsubsection{Dimension Reduction}

To visualize the distribution of all strides (patients and therapists, both legs), we employed dimensionality reduction techniques to map each high-dimensional stride $s\in S$ into a two-dimensional latent space representation $z \in Z \subset \mathbb{R}^2$. Several dimension reduction techniques were evaluated, including Principal Component Analysis, Autoencoders, and VAE. Among these, the VAE was selected due to three key advantages: (i) its ability to construct a compact, normally distributed latent space, (ii) its capacity to accurately reconstruct the original strides, and (iii) its strong generalization ability across unseen data.

The architecture of the VAE is shown in Fig.~\ref{fig:initialImg}. It consists of an encoder ($f:\mathbb{R}^{100}\rightarrow \mathbb{R}^2$) and a decoder ($f^{-1}:\mathbb{R}^2\rightarrow \mathbb{R}^{100}$). The encoder maps a high-dimensional stride to a compact latent representation, while the decoder inverts this mapping to reconstruct the full stride from the latent space. The VAE is trained autoregressively by comparing input strides (processed in batches) with their reconstructions, using the mean squared error as the cost function.

The encoder takes as input a stride $s$ and processes it through two successive 1D convolutional layers interleaved with max-pooling layers, followed by a stack of dense layers. These layers output the parameters of the latent distribution, namely the mean ($z_{\mu} \in \mathbb{R}^2$) and standard deviation ($z_{\sigma} \in \mathbb{R}^2$), which are used to sample the two-dimensional latent representation $z$. The decoder then reconstructs the stride by expanding the low-dimensional latent space through a series of dense layers. The final layer generates $(a, b) \in \mathbb{R}^n$ coefficients used to weight a Fourier basis function 
\begin{equation}
    \hat{s}(\tau) = \sum_{k=1}^{n/2} \Big( a_k \cos( \frac{2\pi k\tau}{100}) + b_k \sin(\frac{2\pi k\tau}{100}) \Big), 
 \tau \in [0,100].
\end{equation}

This Fourier-based reconstruction ensures that the reconstructed stride is smooth and continuous, faithfully preserving the cyclic nature of the gait.
Each joint (hip and knee) was modeled independently, resulting in two separate VAEs. A preliminary optimization phase was conducted to fine-tune the parameters of each VAE by investigating the joints individually. The final set of parameters used for training is summarized in Fig~\ref{fig:initialImg}. Each model was pre-trained using a previously collected dataset \cite{lhoste2025benchmark} and then trained for 1000 epochs with a batch size of 256, using an early stopping criterion with a patience of 20 epochs to prevent overfitting and ensure generalization.

\subsubsection{Force Field Creation}

The encoder previously trained was used to reduce the dimensionality of the stride dataset $S$, yielding the latent space representation $Z = f(S)$. This latent space was then employed to compute the virtual interaction forces between patient and therapist as:
\begin{align}
    F_p &= K_p \,(z_{t} - z_{p}), \\
    F_t &= K_t \,(z_{p} - z_{t}),
\end{align}
where $z_{t}$ and $z_{p}$ denote the latent space representations of the therapist’s and patient’s kinematics, respectively, and $K_p$ and $K_t$ are the stiffness parameters associated with the patient’s and therapist’s exoskeletons.

Each force field was calculated independently for each joint. The resulting dataset ($Z, F_i$), comprising the latent space representations of joint kinematics and the corresponding virtual forces, was split into training and validation sets. A Gaussian Mixture Model (GMM) was then used to learn the joint distribution:

\begin{equation}
p(z_{p}, F) = \sum_{k=1}^K \pi_k , \mathcal{N}\big([z_p, F]^T \mid \mu_k, \Sigma_k \big),
\end{equation}

where $K$ denotes the number of Gaussian components used to fit the model. The parameters were optimized on the validation set to prevent overfitting, and the final model employed $K=10$.

Once the GMM has been trained, it can be used to reconstruct the full force field in the latent space as displayed in Fig. \ref{fig:LatentAndForceFields}. First, a set of points $Z_s$ is sampled uniformly within the latent space, spanning the range defined by the minimum and maximum values of the latent variables $Z_s \sim U(\min(z), \max(z))$. 
These sampled latent points are then passed through the decoder to reconstruct the corresponding strides ($S_s = f^{-1}(Z_s)$). Finally, for each latent point, the expected interaction force is inferred from the trained GMM as $F = \mathbb{E}[F \mid z]$. 
The combination of the reconstructed strides $S_s$ and their associated forces $F$ across the sampled latent space allows the visualization of the continuous force field. This approach provides a comprehensive representation of the therapist-patient interaction dynamics, capturing how corrective forces vary across different regions of the latent space and enabling subsequent analysis or simulation of patient-specific rehabilitation strategies.

%% file: sections/05_SyntheticTherapistMethods.tex
\subsection{Synthetic Therapist}
In this section, we present our approach to creating the ST for post-stroke rehabilitation using previously collected patient–therapist interaction data. We employ a Long Short-Term Memory (LSTM) model to capture long-term dependencies between patient and therapist kinematics. The approach was validated using leave-one-out cross-validation across eight post-stroke patients and integrated into the exoskeleton controller via the Robot Operating System (ROS). We first describe the data processing, then the model design, training, validation, and ROS integration.

\subsubsection{Data Processing} 

The dataset presented in Sec. \ref{sec:dataset} has been temporally aligned between the patient and therapist.
Across all features present in the dataset, only the following were retained: patient joint positions ($q_p \in \mathbb{R}^4$) and velocities ($\dot{q}_p \in \mathbb{R}^4$), and therapist joint positions ($q_t \in \mathbb{R}^4$) and velocities ($\dot{q}_t \in \mathbb{R}^4$). We decided to focus on these features after comparison with different models including additional parameters (e.g., weight distribution $\alpha \in \mathbb{R}$ and interaction torque $\tau_{int} \in \mathbb{R}^4$). Incorporating additional features could potentially provide more information about the patient’s kinematics and dynamics; however, it may also introduce additional noise due to the limitations of the sensors used for data collection, 
\begin{color}{black}
namely force-sensing resistors (FSRs)
\end{color}
and strain gauges. Collecting more data could help mitigate the impact of this noise.

The temporal data were segmented into sliding windows of constant length, each consisting of 50 timesteps, corresponding to approximately $150 \text{ ms}$ in the time domain. Each window was shifted by a step of 10 timesteps (i.e., $30 \text{ ms}$). The choice of a sliding window of dimension $150 \text{ ms}$ reflects the trade-off between enabling fast inference and preserving sufficient data to capture relevant features, while focusing exclusively on the kinematics of the current step and avoiding reliance on information from previous strides.
The ST model was trained using a memory window which half overlapped with the incoming data, enabling the learning of patient gait patterns, and half extended beyond the incoming data, allowing for future predictions. Specifically, given patient kinematics over the interval $[t - T, t]$ with $T = 150$ ms, the ST infers the therapist kinematics for the window $[t - \frac{T}{2}, t + \frac{T}{2}]$. The exact time point used for inference can vary depending on the application and training requirements. For instance, the ST may provide only a guiding response by using the prediction at time $t$, or it may deliver proactive assistance by incorporating predictions at future time steps. This flexibility allows therapists to tailor the degree of robotic assistance to the patient’s needs and the therapeutic objectives.
Finally, for the model training–validation–testing pipeline, the data were divided following a 70–20–10 split for each patient-therapist dyad individually.



\subsubsection{Model Structure and Training} 

Eight independent LSTM models were used to predict each therapist feature individually ($q_t$, $\dot{q}_t$).  Each model was implemented in PyTorch and receives the eight patient features ($q_p$, $\dot{q}_p$) within a sliding window of length 50. The resulting input tensor had dimensions $(50, 8)$. The input was first processed by the LSTM and subsequently passed to a linear layer, as illustrated in Fig.~\ref{fig:initialImg}. The output of the model is the future therapist kinematics for a specific feature (50, 1). 
The model parameters were optimized through a preliminary heuristic search, while the decision to train each feature independently was motivated by the aim of reducing output correlation and was validated in preliminary experiments.

Each model was trained independently for 150 epochs, with validation performed at each epoch, using a batch size of 256, shuffled samples, and a learning rate of $1\times 10^{-5}$ within the Adam optimizer. 
The eight trained models were then conglomerated into one and scripted for performance optimization and deployment. The combined model receives the full patient kinematics (50, 8) and returns the therapist kinematics for the time window (50, 8). 

The ST was validated using a leave-one-out approach in which repeated iterations of the model were trained on seven of the eight patients and tested on the remaining patient. This approach has been used to investigate the ability of the model to generalize to unseen data collected in different experimental conditions.
For leave-one-out, the window maintained the same size as the ST model, 50 time steps, with the halfway point being the next timestep after the patient input data. When calculating the root Mean Square Error (rMSE) of the ST predictions compared to true therapist data, the entire window is used to create one data point, meaning the final rMSE contains information from all windows across all timesteps.


\subsubsection{ROS Integration}
The model was integrated into the current exoskeleton controller to validate inference time and support a preliminary implementation of future training protocols. Specifically, we developed a ROS-based integration of the ST, which receives patient kinematics from the exoskeleton and infers the corresponding therapist behavior. A ring buffer with a maximum length of 50 (matching the sliding window) is used to store incoming patient exoskeleton joint kinematics while the node runs model inferences at a rate of 333 Hz to match the frequency of the incoming data. The ROS integration was validated both on offline collected data and on real-time streaming data from the exoskeleton. 


%% file: sections/06_ForceFieldResults.tex
The VAE reconstructs the joint kinematics of the overall dataset with an rMSE of $4.54 \pm 1.65$ degrees (mean rMSE $\pm$ std) for the hip and $4.20 \pm 2.14$ for the knee. These results vary substantially depending on the source of the stride. For the patients, the overall rMSE is $4.57 \pm 1.73$, while for the therapists it is $4.67 \pm 1.68$. The variability is even larger when examining each individual patient and comparing the left and right legs.
For instance, patient 4 presents an rMSE of $3.18 \pm 0.95$ for the left leg and $5.09 \pm 1.44$ for the right leg, while the corresponding legs for the therapist are $3.58 \pm 1.14$ and $4.76 \pm 1.10$, respectively. These discrepancies can be attributed to how different the patients’ strides are compared to the mean stride profile and may serve as an indicator of impairment level. At the same time, the therapist attempts to overcompensate for the patients’ kinematics, which also results in abnormal gait patterns.

While the latent features have been automatically selected by the VAE, their representation reveals several interesting properties. First, the distances between the latent positions of the left and right legs for some patients, as shown in Fig.~\ref{fig:LatentAndForceFields}.Top, highlight the asymmetry of their gait patterns. For instance, patient 4 exhibits a distance more than three times greater than that of patient 5.
Second, the $z_1$ coordinate encodes a shift in the timing of knee flexion, while $z_2$ represents its amplitude, suggesting that these two metrics may serve as meaningful latent features for representing gait patterns.
Finally, although the same therapist performed training with all patients, the therapist’s latent profile differs depending on the patient. For example, with patient 1 the latent representation of the therapist’s kinematics lies in the upper-right hemisphere of the latent space, while with patient 4 it is located in the opposite hemisphere. This result suggests that therapists adapt their kinematics rather than simply performing their natural gait, adjusting to the specific needs of each patient—and doing so separately for each leg.

In Fig.~\ref{fig:LatentAndForceFields}, the two force fields are shown: the patient \begin{color}{black}field\end{color} with virtual force $F_p$ and the therapist \begin{color}{black}field\end{color} with virtual force $F_t$, for both joints (hip and knee). The GMM fits the distribution of the force fields with a log-likelihood of $-4.4 \pm 2.15$ for the hip and $-4.75 \pm 1.90$ for the knee. From the patient force field plots, we can observe how some regions act as attractors for the patient kinematics. Namely, the therapist attempts to drive the patient toward specific regions of kinematics that are considered more clinically relevant. Conversely, the therapist's force field suggests how the haptic feedback perceived by the therapist is pushed away from those same regions.

While, at first glance, the two force fields may appear symmetric (i.e., $F_t = -F_p$), in many experimental scenarios, the therapist's stiffness is a rescaled (reduced) version of the patient's stiffness. These stiffness parameters are chosen by the therapist during training, and the visualization provides a first insight into how such parameters are selected.
One hypothesis is that the therapist may intentionally modify the shape of “unstable regions” for themselves while walking, while simultaneously desiring strong haptic feedback in other clinically relevant regions. Although these plots represent only a first step toward visualizing such behaviors, they may serve as a foundation for future approaches aimed at designing adaptive tuning strategies for therapist stiffness.

%% file: sections/07_SynthticTherapistResults.tex
Tab. \ref{tab:leaveOneOut} displays the results of the leave-one-out approach. To summarize the data for each of the eight models,  we report the mean of the four position and velocity rMSEs.

Each model in the leave-one-out approach predicted the unseen patient data with various levels of accuracy, potentially resulting from patient impairment level, patient-therapist coordination, or patient range of motion. PT$_{5}$ has the lowest rMSE for both joint position and velocity, indicating that the data collected from that patient is more similar to the training set compared to others like PT$_{3}$, with the worst performance metrics for both position and velocity \begin{color}{black}despite the same therapist being used throughout data collection\end{color}. Regardless, the range of error from the minimum to maximum rMSE is less than 3 degrees for position and just over 15 degrees per second for velocity. This results in $5.05 \pm 0.85$deg mean rMSE for position and $23.19 \pm 2.64$deg/sec for velocity across all leave one out models.

The ST predicts therapist reactions to patient input data with $4.30 \pm 0.41$deg for joint position and $20.67 \pm 1.47$deg/s for joint velocity.
The ST model, trained on all patient datasets and tested on a conglomeration of all patients testing data (previously unseen by the model), shows a lower mean rMSE for both joint position and velocity than most of the leave-one-out models, indicating that it can generalize across patients.

Fig. \ref{fig:PredVsTrue} displays an example of ST prediction across a gait pattern for the hip and knee joints. Similar results were observed across all patients. For the hip joint (top figure), the largest deviations between the model predictions and the true kinematic data occur during stance, a phase which is more variable due to patients reacting differently to ground interaction and weight shifting. Additionaly, noise from the FSRs sensors during weight shifting may introduce oscillatory artifacts in the stance leg, an effect that is more pronounced in the hip joint: hip predictions tend to deviate more from the target values compared to the knee joint (bottom figure), likely due to the lower inertia of the knee, which reduces oscillatory behavior.

The ST was integrated into a ROS node for real-time prediction of joint kinematics, enabling future validation during actual training sessions. Using an Intel Core i9-14900HX CPU, the model achieved an average inference time of less than 3 milliseconds, which is within the desired range and supports a streaming frequency of 333 Hz for real-time implementation.

%% file: sections/08_Conclusions.tex
In this work, we propose two novel approaches:
1) A \textit{ Patient-Therapist Force Field} to visualize the therapist’s response to patient kinematics, providing a better understanding of therapist policy. The method combines a Variational Autoencoder to learn latent features representing patient and therapist joint kinematics with a Gaussian Mixture Regression to model the force field attracting these features between therapist and patient. This approach can potentially reveal therapist strategies and inform the design of novel rehabilitation robot controllers.

2) A \textit{Synthetic Therapist} that learns from patient–therapist demonstrations to infer therapist responses using an LSTM and a sliding time window. The ST predicts therapist responses up to 75 ms into the future with less than 3 ms inference time and has been integrated into a ROS node, embedding it into the patient exoskeleton control loop to provide desired joint angles and velocities, thus generating interaction torque. Future work involves testing in real-time the proposed approach with the same post-stroke patients present in the training set, as well as novel patients.

Both approaches would benefit from further data collection to better understand generalization capabilities. Additionally, experiments with diverse patient populations and different mechanical connections could clarify the role of connection parameters, namely stiffness and damping.